# Design Project of an Open-Source, Low-Cost, and Lightweight Robotic Manipulator for High School Students

Isabella Huang[*,1] , Qianwen Zhao[*,2], Maxine Fontaine[2], and Long Wang[2]

[1]*Watchung Hills Regional High School, 108 Stirling Road, Warren, NJ 07059*

[2]*Stevens Institute of Technology, 1 Castle Point Terrace, Hoboken, NJ 07030.*

## Abstract

In recent years, there is an increasing interest in high school robotics extracurriculars such as robotics clubs and robotics competitions. The growing demand is a result of more ubiquitous open-source software and affordable off-the-shelf hardware kits, which significantly help lower the barrier for entry-level robotics hobbyists.

In this project, we present an open-source, low-cost, and lightweight robotic manipulator designed and developed by a high school researcher under the guidance of a university faculty and a Ph.D. student. We believe the presented project is suitable for high school robotics research and educational activities. Our open-source package consists of mechanical design models, mechatronics specifications, and software program source codes. The mechanical design models include CAD (Computer Aided Design) files that are ready for prototyping (3D printing technology) and serve as an assembly guide accommodated with a complete bill of materials. Electrical wiring diagrams and low-level controllers are documented in detail as part of the open-source software package.

The educational objective of this project is to enable high school student teams to replicate and build a robotic manipulator. The engineering experience that high school students acquire in the proposed project is full-stack, including mechanical design, mechatronics, and programming. The project significantly enriches their hands-on engineering experience in a project-based environment. Throughout this project, we discovered that the high school researcher was able to apply multidisciplinary knowledge from K-12 STEM courses to build the robotic manipulator. The researcher was able to go through a system engineering design and development process and obtain skills to use professional engineering tools including SolidWorks and Arduino microcontrollers.

## Introduction

Science Technology Engineering Mathematics (STEM) education promotes critical thinking, innovation, and problem-solving, which has become critically important for the future of students and our national economy [1, 2, 3]. The US high school curriculum has many courses dedicated

* Equal contribution.

Q. Zhao, M. Fontaine, and L. Wang are with the Department of Mechanical Engineering, Stevens Institute of Technology, New Jersey, NJ, 07030, USA, e-mail: lwang4@stevens.edu

This work has been supported in part by National Science Foundation grant #CMMI-2138896.

to this area to inspire young students to explore and experiment. However, these courses are more theoretical and conceptual on their own subjects, thus there is a lack of interdisciplinary approaches to link all subjects together. Robotics is an inherently interdisciplinary field that brings together math, physics, cognitive science, computer science, electrical engineering, and mechanical engineering. A robotics education in high school fulfills this acute need of preparing students to engage in diverse fundamental STEM concepts, in math, physics, engineering, computer programing, and industry design [4,5].

The current challenge to implementing robotics curriculums in K-12 education is the lack of suitable hands-on projects for starters. In this project, we designed and developed an open-source, low-cost, and lightweight robot manipulator that can be easily adopted and replicated by high school students who are interested in exploring and learning engineering and robotics in college.

Through the particular project described in this work, in addition to the gain on theoretical and conceptual knowledge such as math, physics, mechanical engineering, a high school student will obtain practical hands-on skills such as using CAD software and microcontroller programming, which can benefit their other robotics club and hobbyist activities.

## Related Commercialized Products

In the current market, there are a few commercially available manipulator kits that are modeled for gripping. Some advanced models are equipped with artificial intelligence and machine learning. Below we analyzed a couple of existing manipulator models on the internet.

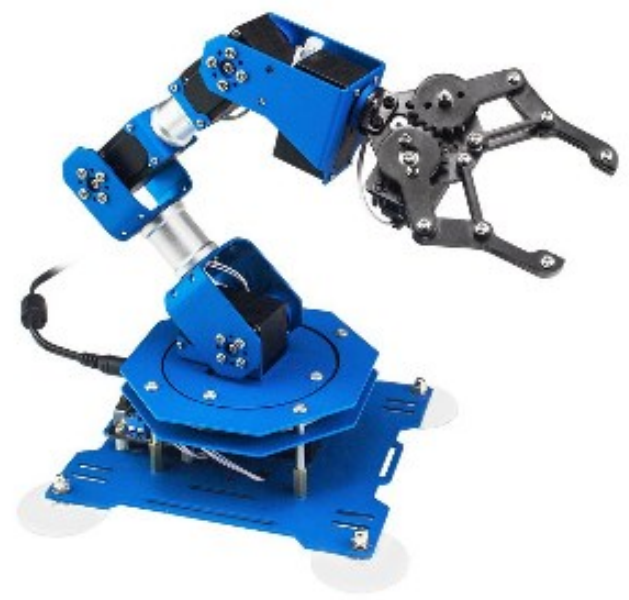

Figure 1: Robotic xArm 6-DOF Arm

1) xArm [4] - This robotic manipulator model (Fig. 1) is designed with a wide square base, four vacuum suction cups to fix it on a table, and six degrees of freedom (five servos and one digital servo). The three joints of this arm are composed of metal brackets attached to both sides of the motor to maintain balance, and the gripper is a thin mechanical claw that can extend and contract to pick up small items on the table. The metal components and wide base of this model generate a heavier manipulator which would be difficult for transportation.

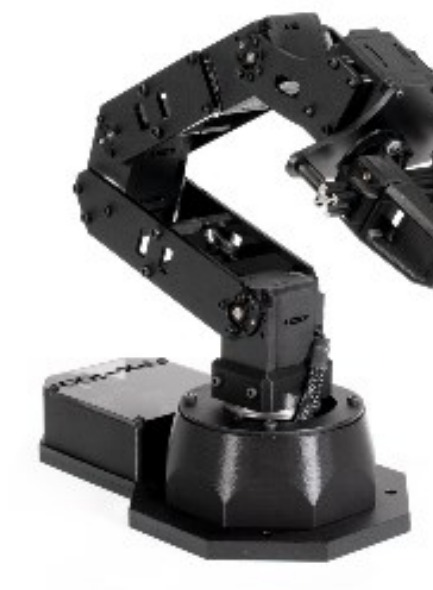

Figure 2: PincherX 100 Robot Arm

2) Trossen Robotics PincherX [5] - This robotic arm (Fig. 2) leans on the expensive side ($650) as it is equipped with multiple machine learning and artificial intelligence algorithms targeting classroom functions. It has four degrees of freedom and a seventy-five-centimeter horizontal reach. A specialty in the PincherX design is its small and lightweight base which can easily be adapted to drones.

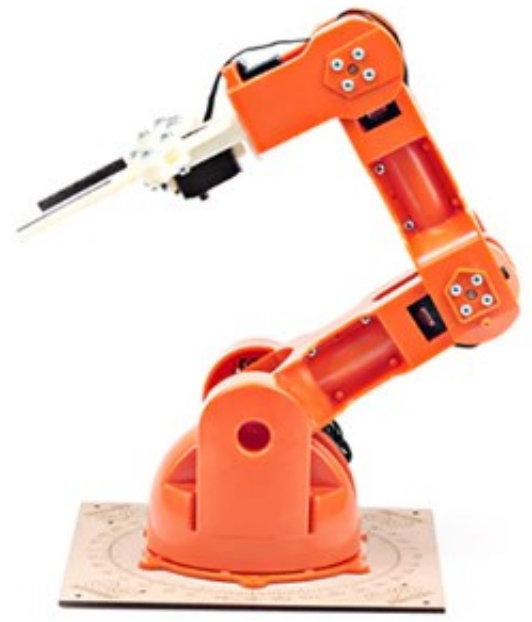

Figure 3: Tinkerkit Braccio Robot

3) Tinkerkit Braccio Robot [6] - Similar to the small base design of the Trossen Robotics PincherX, the Tinkerkit Braccio robot (Fig. 3) has a spherical base that is able to rotate from left to right. This lightweight and transportational robot is composed of six servo motors, three joints, and a mechanical claw. Although the mechanical claw has a wider reach, its limited adaptability reduces the variety of objects it is able to pick up. This Tinkerkit Braccio robot also has the ability to be paired with solar panels which can help lengthen the battery life of the manipulator.

In this paper, we present a robotic manipulator with 5-DOF emphasizing its lightweight character and maneuverability. We also take into consideration the cost of materials as well as the construction ability of K-12 education.

## Project Requirements

This work provides high school students with an open-source robot design project that uses easily accessible and affordable materials. Five servo motors (Dynamixel XL430, or a smaller version Dynamixel XL330) were selected as the robot actuators. An Arduino Mega and Dynamixel motor shield function as the main controller of the robot. The Dynamixel motor shield is an add-on board to the Arduino Mega (or an Arduino Uno) that can be directly plugged in to use. The well-known Arduino IDE serves as the programming interface. The overall cost of the hardware can vary from $170 to $317 (market price by Oct, 2022) depending on the choice of motors and microcontrollers. The link to the open-source Github repository is provided here: https://github.com/stevens-armlab/desk-robot-manipulator .

The prototyping of the designed robotic manipulator is completed through 3D printing. In recent years, 3D printers are widely available in colleges, universities, and online stores. Many high schools also have affordable ones that students may use with the help of lab assistants.

## Mechanism Design and Electronics

A simple hobbyist robot design usually consists of the following:

1) Mechanical designs and prototypes.
2) The software that drives the robot.

For a group project, the hardware and software development can work in parallel. For an individual project, it is recommended to start with the mechanical design.

SolidWorks was selected as the 3D modeling tool since it is a professional CAD software used for educational purposes and in the industry. We designed and 3D printed the following:

1) The robot base.
2) The robot manipulator linkages.
3) The parallel gripper.

The manipulator has five degrees of freedom and a base that can rotate 360° degrees. On top of the foundation sits three joint links to enable the robotic arm's 30 cm horizontal and 40 cm vertical mobility. Each joint is powered by a Dynamixel motor. At the end of the arm is a padded parallel gripper for securely picking up small, unknown objects for transportation. Figure 4 depicts the mechanical design of the robot manipulator.

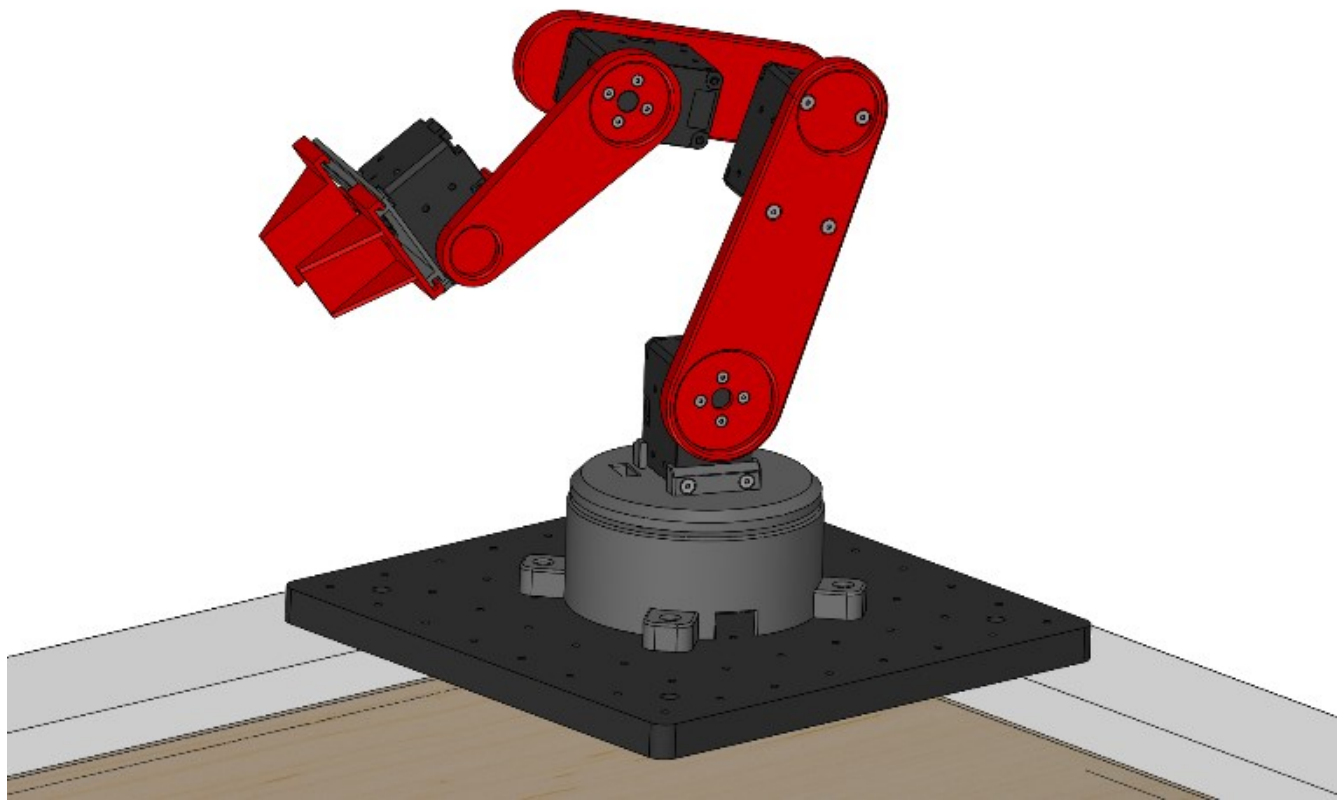

Figure 4: An illustration of the desktop robot manipulator design

The low-level control of the robot manipulator is done through an Arduino microcontroller and an add-on Dynamixel Shield. The five servo motors are chained together and only one connector is needed to be plugged into the control unit. Always being aware of cleanness and robustness when designing a robot (or generally, a product) is a good design tip for future engineering students. The robot is powered by a 12VDC wall-plug power adapter.

The SolidWorks files of parts and the assembly are open-sourced and available for downloading. High school students who are interested in this work can download CAD files (make modifications as needed) and send the corresponding STL files to be 3D printed.

## Mechanical Design Details

The base of the manipulator is a 9 cm hollow cylindrical base (Fig. 5) with four extended 1 cm anchoring tabs. In the hollow area of the base sits a 3 cm thick motor housing strong enough to secure the Dynamixel XL430-W250-T servo motor (Motor 1) without adding excessive weight. A 5-mm plate is mounted directly on this motor to allow 360-degree rotation motion.

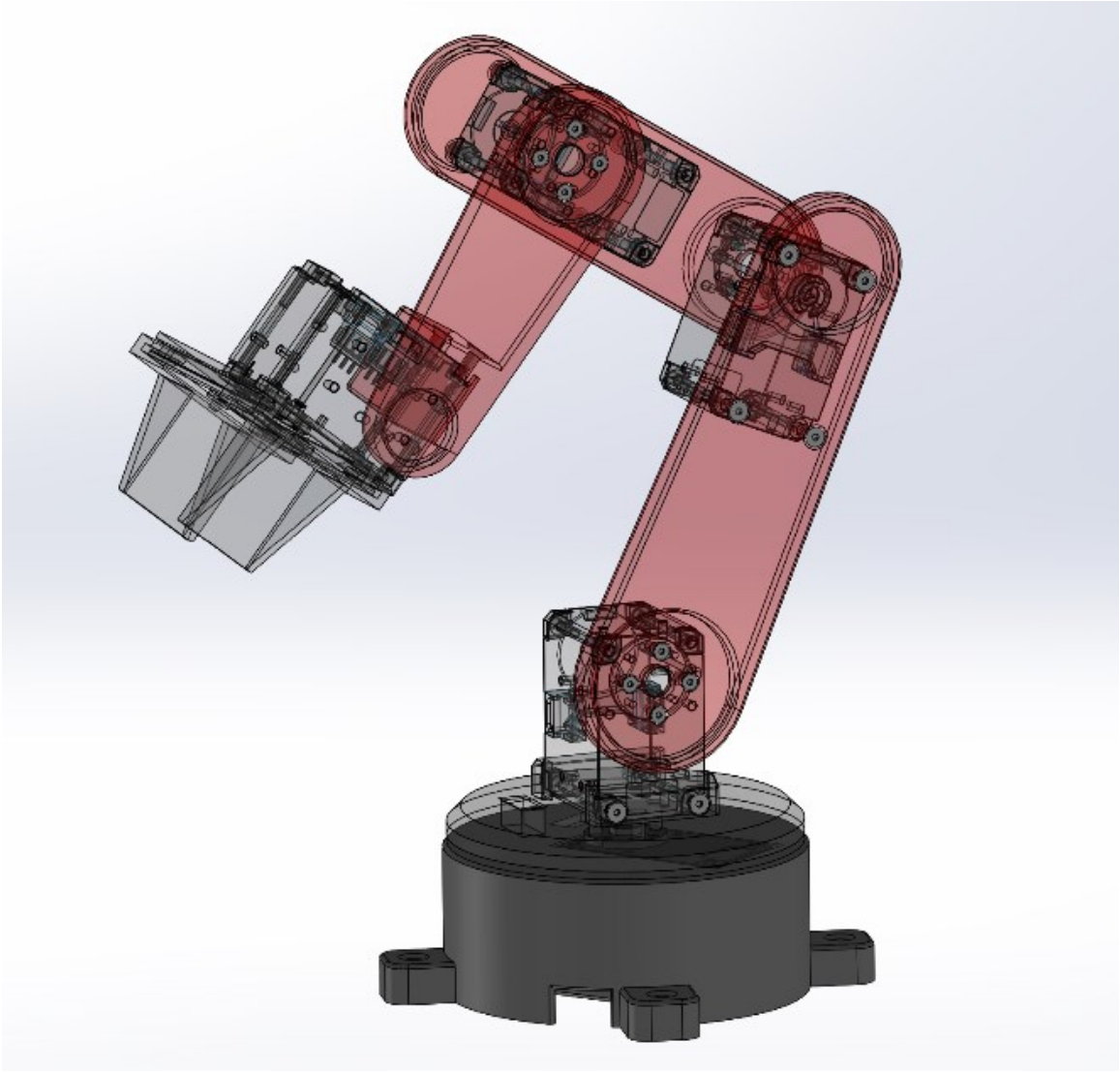

Figure 5: The manipulator design with the base highlighted.

A second motor is mounted on the plate (Fig. 6) through a rectangular prism sitting flush against the motor, holding it in place and supporting the weight of the additional joints.

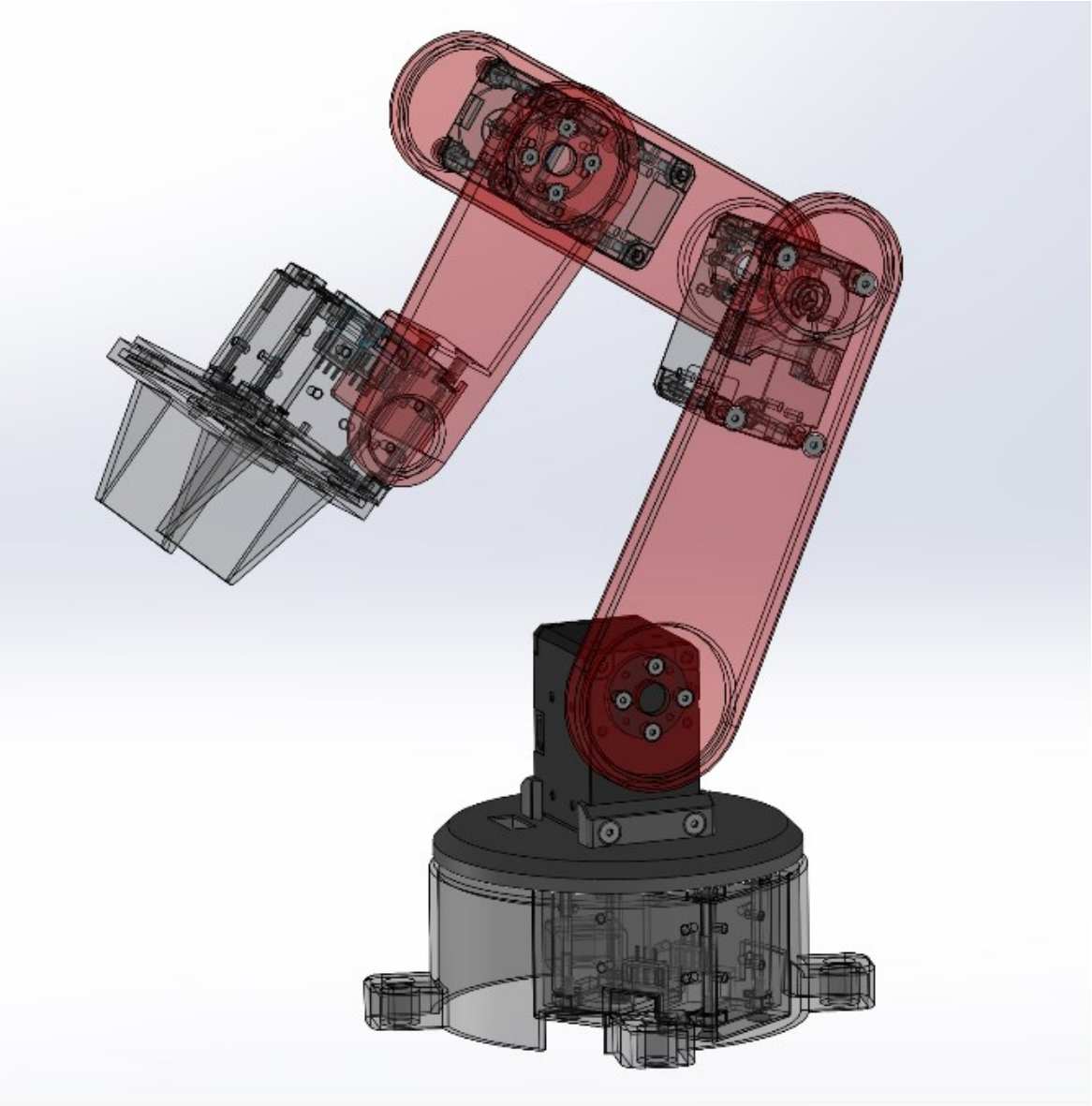

Figure 6: The manipulator design with the second motor and the base plate highlighted.

The second and the third motors are connected by the first arm linkage, as illustrated in Fig. 7. The second motor dominants the motion of the end-effector (gripper) in the vertical plane.

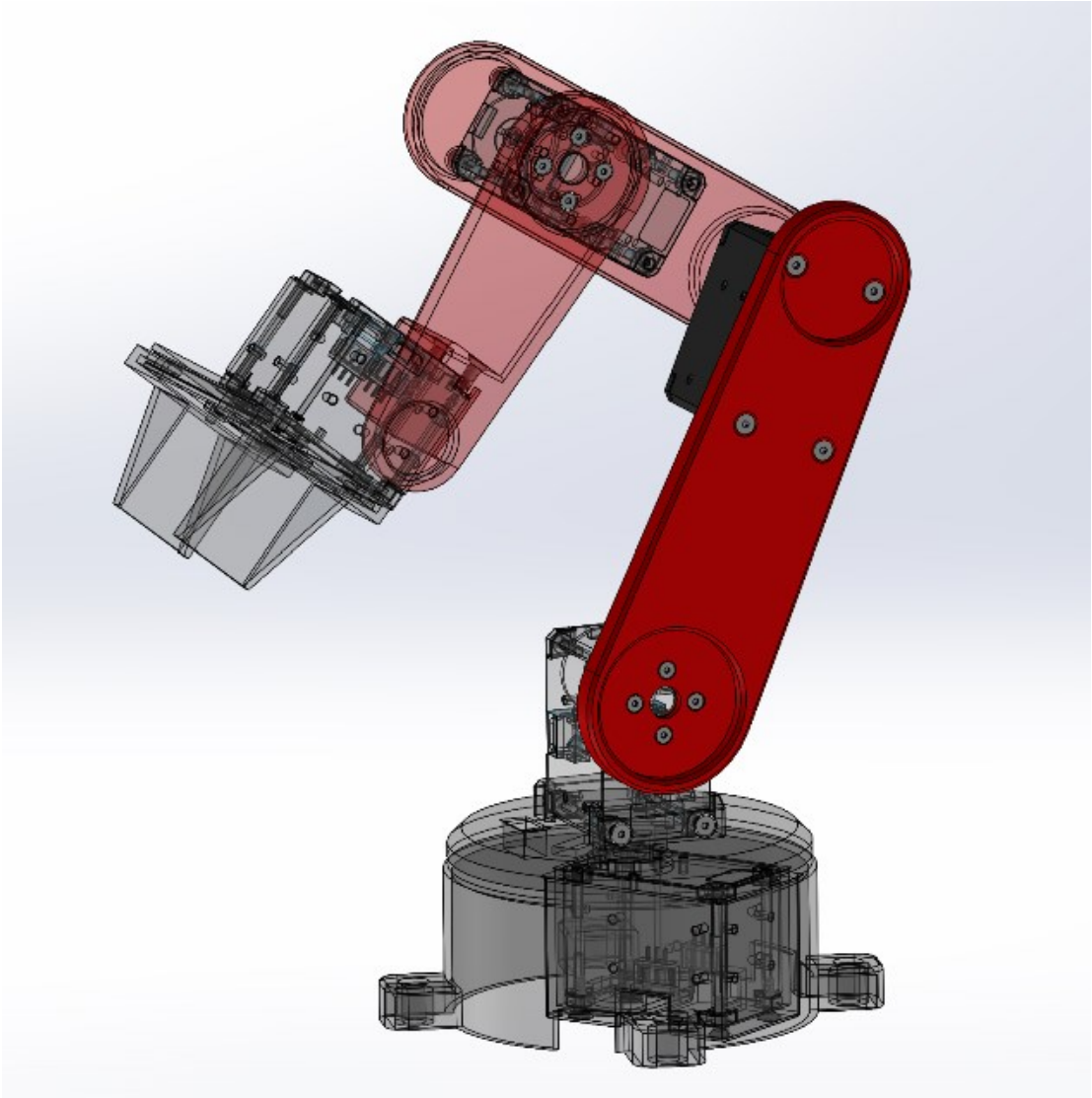

Figure 7: The manipulator design with the third motor and the first arm linkage highlighted.

The third and the fourth motors are connected by the second arm linkages, as shown in Fig. 8. The third motor also contributes to the end-effector's motion in the vertical plane.

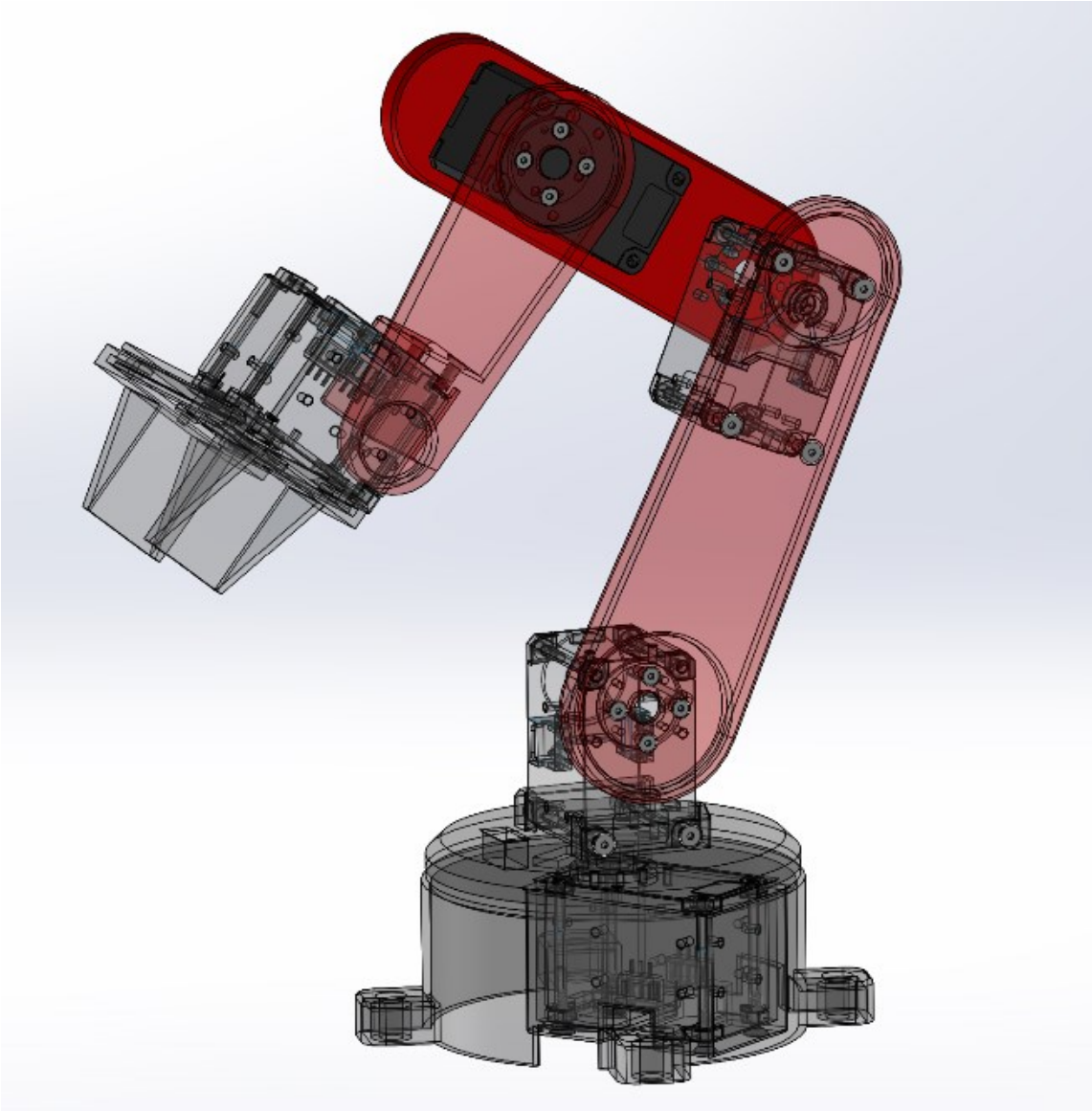

Figure 8: The manipulator design with the fourth motor and the second arm linkage highlighted.

The fourth motor is the last revolute joint of the manipulator, and it also contributes to the end effector's motion in the vertical plane. Fig. 9 shows the last linkage to bridge the fourth motor and the fifth motor, where the fifth motor controls a parallel gripper.

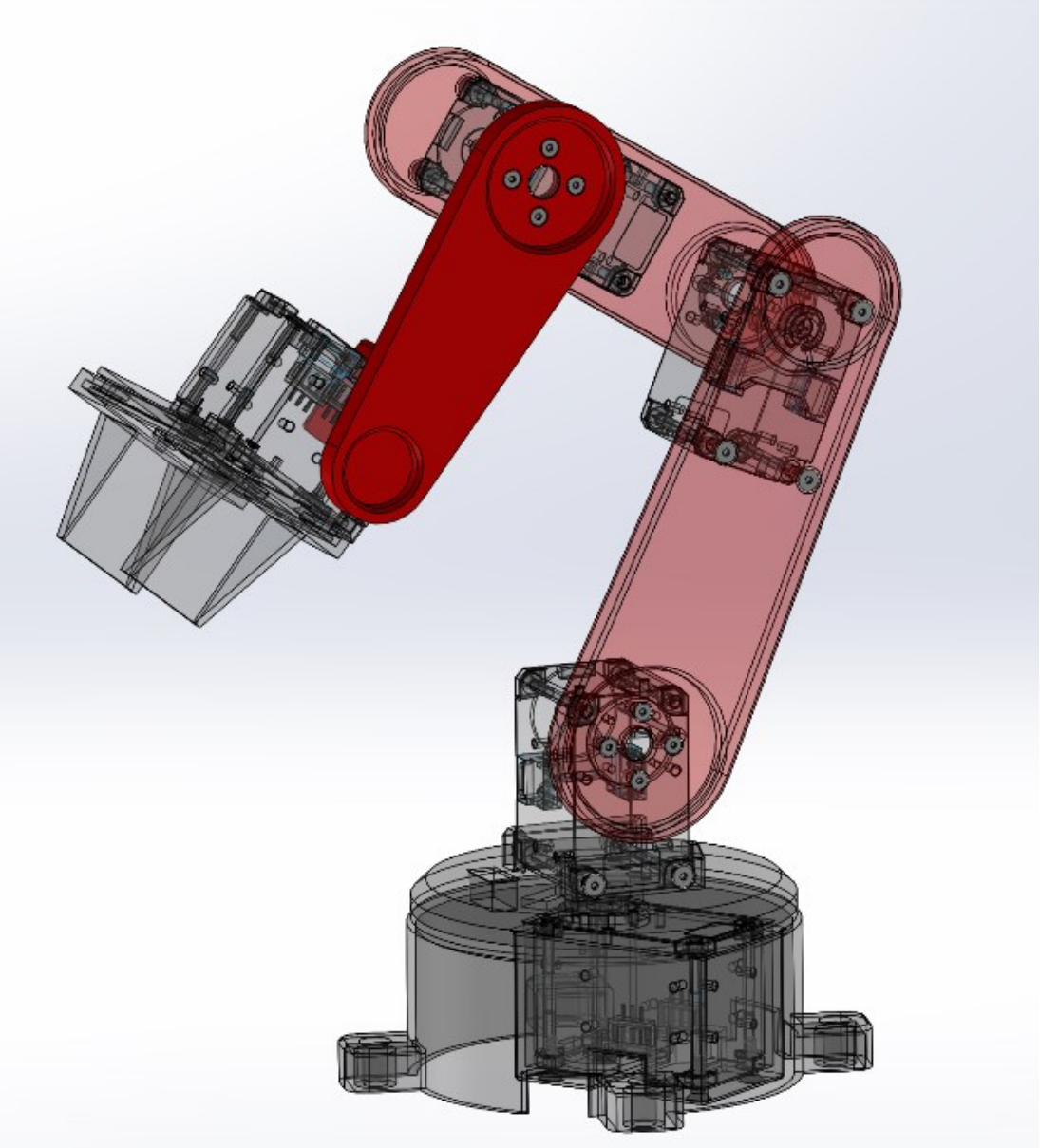

Figure 9: The manipulator design with the fifth motor and the third arm linkage highlighted.

The parallel gripper design is demonstrated in Fig. 10. We chose a parallel gripper to create simplicity, and it has a wider range of object dimensions that we could grip. The gripper's width is driven through the rotation of the Dynamixel motor and is connected by two banana-shaped links.

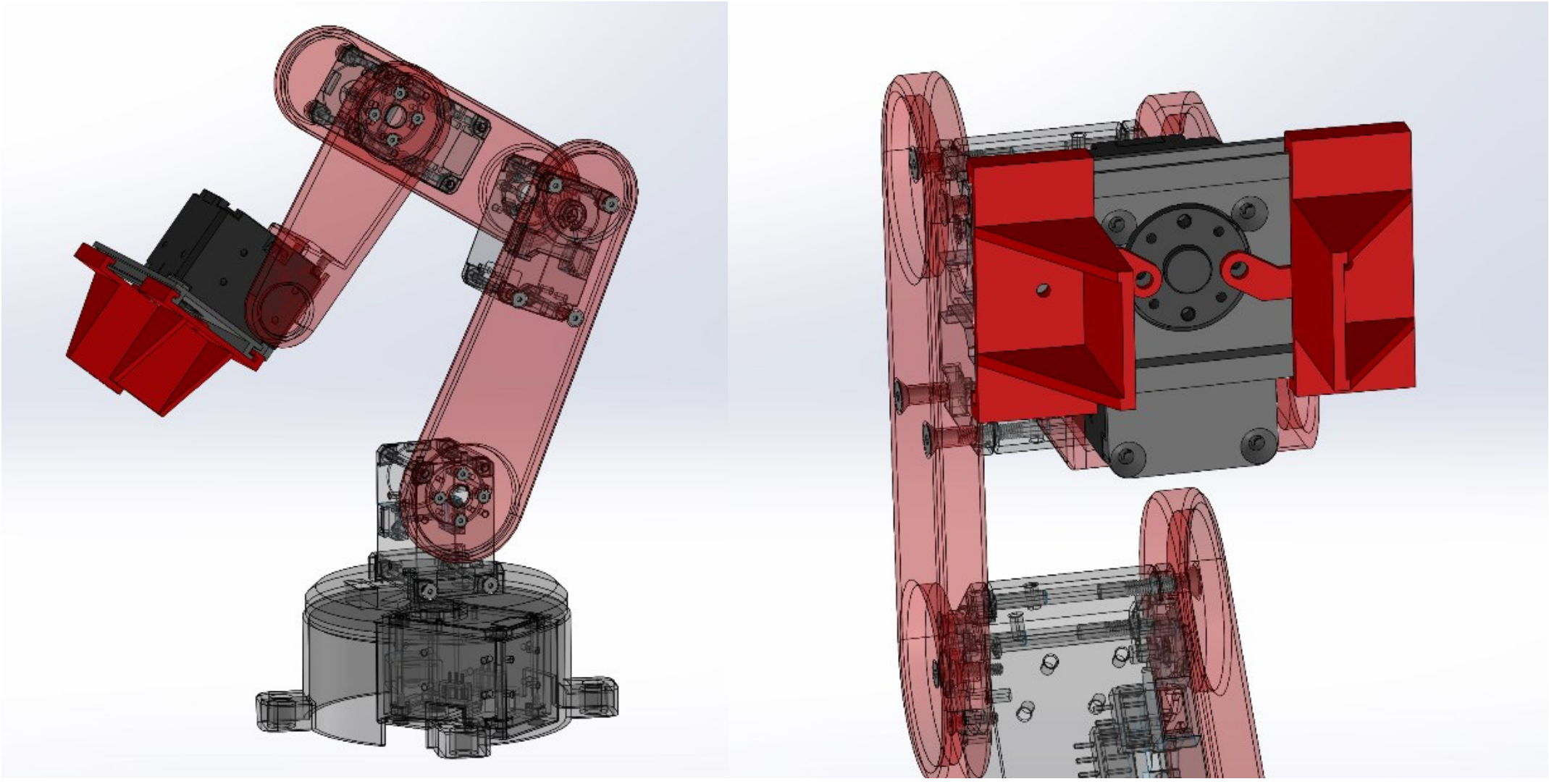

Figure 10: The manipulator design with the gripper highlighted (left), and details on the parallel gripper (right)

An exploded view in included below for reference in assembly and relationships between the different components (Fig. 11).

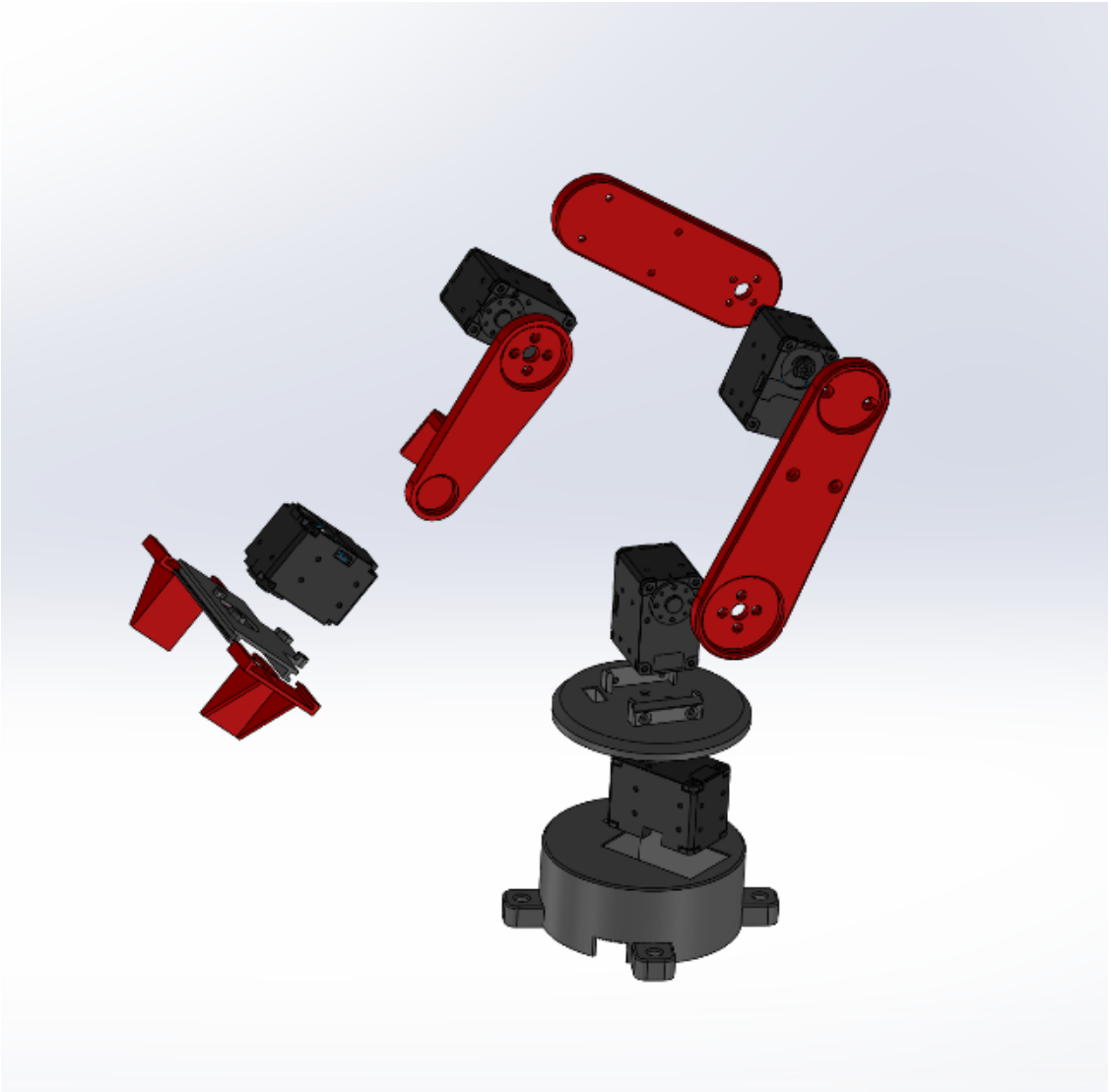

Figure 11: The manipulator's exploded view in CAD.

## The Hardware Prototype of the Manipulator

When constructed and 3D printed, the final robotic manipulator model should resemble the image below (Fig. 12).

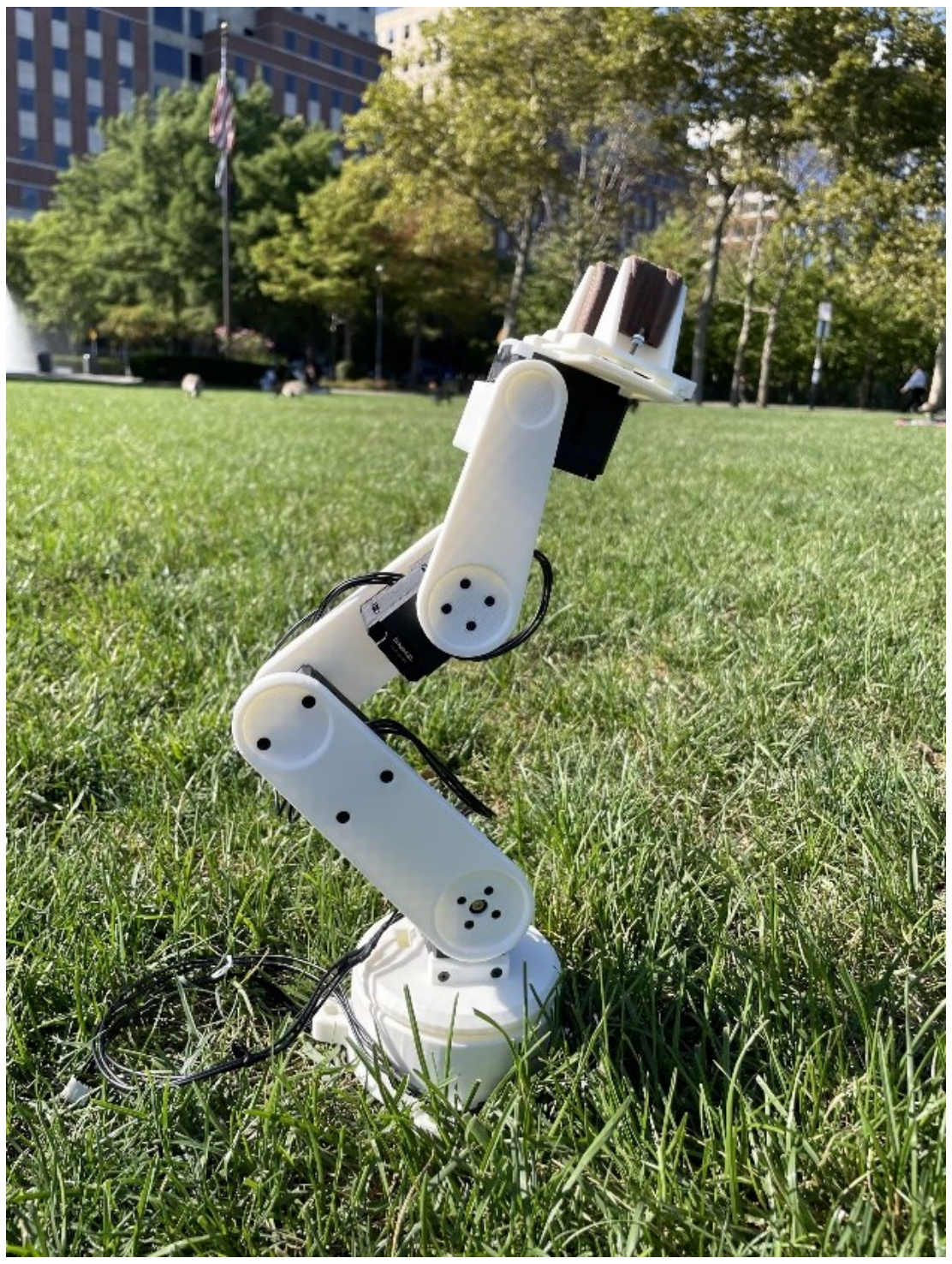

Figure 12: The final robot manipulator prototype

All necessary screws and screwdriver information is included in the download and can be found through McMaster-Carr (https://www.mcmaster.com/)

## Programming

Arduino IDE was used as the programming interface for the desktop robotic manipulator. The Arduino IDE is free to use which makes it the most popular platform for hobbyist programmers. Many manufacturers also develop Arduino libraries for their electronics. We utilized the DynamixelShield Arduino library (https://github.com/ROBOTIS-GIT/DynamixelShield) to program the motion of the robotic manipulator.

Sample code initiating motion of all five servo motors, through keyboard input, is also provided within the open-source package. It uses many Arduino basics such as baud rate, motor identification, and position control to initialize the linkage between each motor and the computer. There are two overall functions which control an increase or decrease in each specified motor. This sample program can be easily modified and advanced after downloading. Such advancements include linking the program with an Xbox controller or using it to grip different objects.

## Project Timeline Management

This desktop robotic manipulator project provides high school students with a full-stack experience in robotics development, ranging from mechanical design and mechatronics integration to software programming. Therefore, the project can either be an individual project or a group project, depending on student preferences and the specific skills that they would like to gain from this project. The following Gantt chart is a typical timeline for a high school student working individually during school days, which means that the student only works on the project 1-2 hours per day.

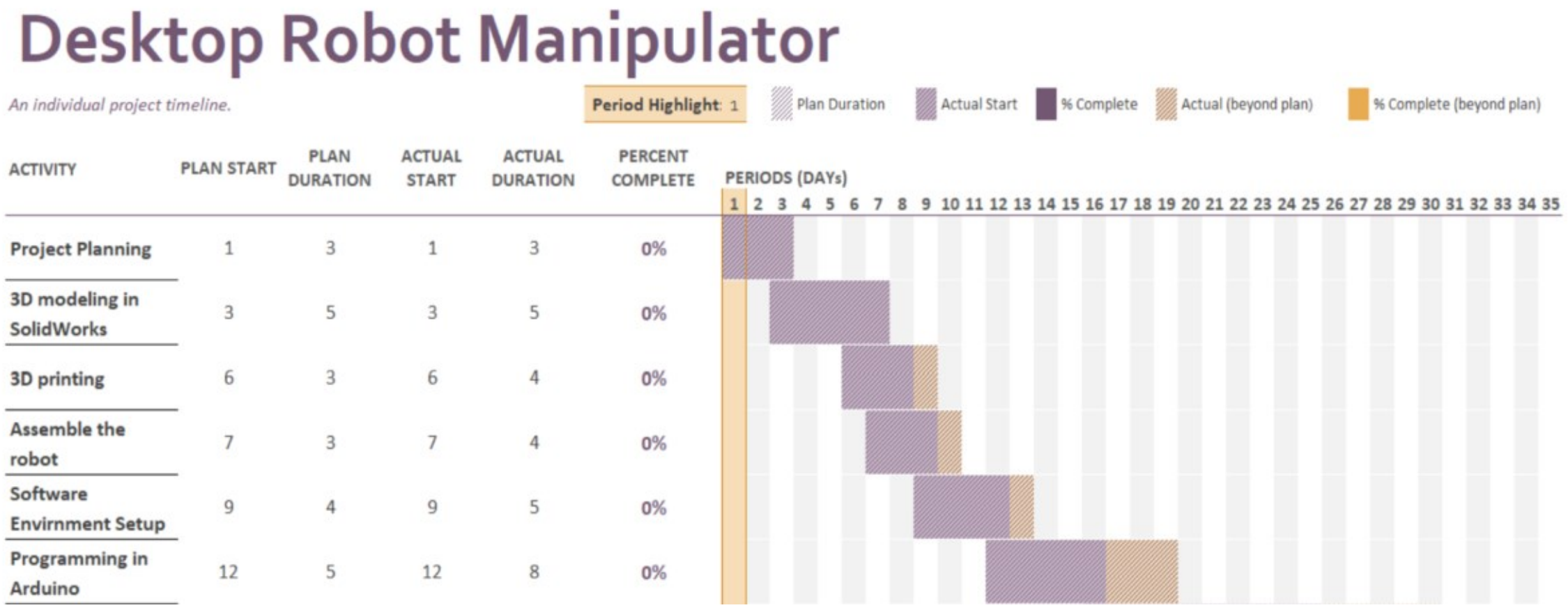

## Conclusion

The project-based design and development environment provides high school students with a great opportunity to apply their learning from K-12 STEM courses to solve real-world problems. Using hobby-level 3D modeling, mechatronics, software programming, and easily accessible materials, this project can be completed independently or added to high school robotics extracurriculars. Through designing and building the open-source desktop robotic manipulator, this experience sparks students' curiosity and creativity and improves their communication, problem-solving, organization, time management, and documentation skills. At the end of the program, students will be familiar with the engineering design process and will gain experience with CAD modeling, basic familiarity with 3D Printing and electronics, and Arduino software programming. They will learn a comprehensive skillset in applying the principles of STEM classes, ultimately promoting their passions beyond high school education.